\documentclass[10pt,twocolumn,letterpaper]{article}

\usepackage[normalem]{ulem}
\usepackage{soul}
\usepackage[pagenumbers]{cvpr} 

\usepackage[dvipsnames, table, svgnames]{xcolor}
\usepackage{graphicx}
\usepackage{amsmath}
\usepackage{amssymb}
\usepackage{booktabs}
\usepackage{subcaption}
\usepackage{caption}
\usepackage{multirow}
\usepackage{tikz}
\usepackage{pgfplots}
\usepackage{fancybox}
\usepackage{calc}
\usepackage{float}
\usetikzlibrary{intersections}
\usepgfplotslibrary{fillbetween}
\usetikzlibrary{spy,calc}
\usepackage{pgfplotstable}
\usetikzlibrary{positioning}
\usetikzlibrary{fit}
\usepackage{algorithm2e}
\usepackage{stmaryrd}
\usepackage[accsupp]{axessibility}

\usetikzlibrary{arrows}
\usepackage{wasysym}

\cornersize{0.1} 

\usepackage{tabularx} 
\newcolumntype{Y}{>{\centering\arraybackslash}X}

\newcommand{\rc}{\rowcolor{gray!10}}
\newcommand{\rb}{\rowcolor{blue!10}}

\usetikzlibrary{arrows,automata}
\usepgfplotslibrary{colormaps} 
\pgfplotsset{compat=1.9}
\pgfplotsset{
colormap={bright}{
	rgb255=(130,180,110) 
    rgb255=(200,80,255) 
	rgb255=(98,70,100) 
    rgb255=(255,113,26) 
    rgb255=(130,160,235) 
    rgb255=(153, 204, 255)
    rgb255=(230,255,0)
    rgb255=(255,255,255)}
}
\newif\ifblackandwhitecycle
\gdef\patternnumber{0}

\pgfkeys{/tikz/.cd,
    zoombox paths/.style={
        draw=orange,
        very thick
    },
    black and white/.is choice,
    black and white/.default=static,
    black and white/static/.style={ 
        draw=white,   
        zoombox paths/.append style={
            draw=white,
            postaction={
                draw=black,
                loosely dashed
            }
        }
    },
    black and white/static/.code={
        \gdef\patternnumber{1}
    },
    black and white/cycle/.code={
        \blackandwhitecycletrue
        \gdef\patternnumber{1}
    },
    black and white pattern/.is choice,
    black and white pattern/0/.style={},
    black and white pattern/1/.style={    
            draw=white,
            postaction={
                draw=black,
                dash pattern=on 2pt off 2pt
            }
    },
    black and white pattern/2/.style={    
            draw=white,
            postaction={
                draw=black,
                dash pattern=on 4pt off 4pt
            }
    },
    black and white pattern/3/.style={    
            draw=white,
            postaction={
                draw=black,
                dash pattern=on 4pt off 4pt on 1pt off 4pt
            }
    },
    black and white pattern/4/.style={    
            draw=white,
            postaction={
                draw=black,
                dash pattern=on 4pt off 2pt on 2 pt off 2pt on 2 pt off 2pt
            }
    },
    zoomboxarray inner gap/.initial=5pt,
    zoomboxarray columns/.initial=2,
    zoomboxarray rows/.initial=2,
    subfigurename/.initial={},
    figurename/.initial={zoombox},
    zoomboxarray/.style={
        execute at begin picture={
            \begin{scope}[
                spy using outlines={%
                    zoombox paths,
                    width=\imagewidth / \pgfkeysvalueof{/tikz/zoomboxarray columns} - (\pgfkeysvalueof{/tikz/zoomboxarray columns} - 1) / \pgfkeysvalueof{/tikz/zoomboxarray columns} * \pgfkeysvalueof{/tikz/zoomboxarray inner gap} -\pgflinewidth,
                    height=\imageheight / \pgfkeysvalueof{/tikz/zoomboxarray rows} - (\pgfkeysvalueof{/tikz/zoomboxarray rows} - 1) / \pgfkeysvalueof{/tikz/zoomboxarray rows} * \pgfkeysvalueof{/tikz/zoomboxarray inner gap}-\pgflinewidth,
                    magnification=3,
                    every spy on node/.style={
                        zoombox paths
                    },
                    every spy in node/.style={
                        zoombox paths
                    }
                }
            ]
        },
        execute at end picture={
            \end{scope}
            \node at (image.north) [anchor=north,inner sep=0pt] {\subcaptionbox{\label{\pgfkeysvalueof{/tikz/figurename}-image}}{\phantomimage}};
            \node at (zoomboxes container.north) [anchor=north,inner sep=0pt] {\subcaptionbox{\label{\pgfkeysvalueof{/tikz/figurename}-zoom}}{\phantomimage}};
     \gdef\patternnumber{0}
        },
        spymargin/.initial=0.5em,
        zoomboxes xshift/.initial=1,
        zoomboxes right/.code=\pgfkeys{/tikz/zoomboxes xshift=1},
        zoomboxes left/.code=\pgfkeys{/tikz/zoomboxes xshift=-1},
        zoomboxes yshift/.initial=0,
        zoomboxes above/.code={
            \pgfkeys{/tikz/zoomboxes yshift=1},
            \pgfkeys{/tikz/zoomboxes xshift=0}
        },
        zoomboxes below/.code={
            \pgfkeys{/tikz/zoomboxes yshift=-1},
            \pgfkeys{/tikz/zoomboxes xshift=0}
        },
        caption margin/.initial=4ex,
    },
    adjust caption spacing/.code={},
    image container/.style={
        inner sep=0pt,
        at=(image.north),
        anchor=north,
        adjust caption spacing
    },
    zoomboxes container/.style={
        inner sep=0pt,
        at=(image.north),
        anchor=north,
        name=zoomboxes container,
        xshift=\pgfkeysvalueof{/tikz/zoomboxes xshift}*(\imagewidth+\pgfkeysvalueof{/tikz/spymargin}),
        yshift=\pgfkeysvalueof{/tikz/zoomboxes yshift}*(\imageheight+\pgfkeysvalueof{/tikz/spymargin}+\pgfkeysvalueof{/tikz/caption margin}),
        adjust caption spacing
    },
    calculate dimensions/.code={
        \pgfpointdiff{\pgfpointanchor{image}{south west} }{\pgfpointanchor{image}{north east} }
        \pgfgetlastxy{\imagewidth}{\imageheight}
        \global\let\imagewidth=\imagewidth
        \global\let\imageheight=\imageheight
        \gdef\columncount{1}
        \gdef\rowcount{1}
        
    },
    image node/.style={
        inner sep=0pt,
        name=image,
        anchor=south west,
        append after command={
            [calculate dimensions]
            node [image container,subfigurename=\pgfkeysvalueof{/tikz/figurename}-image] {\phantomimage}
            node [zoomboxes container,subfigurename=\pgfkeysvalueof{/tikz/figurename}-zoom] {\phantomimage}
        }
    },
    color code/.style={
        zoombox paths/.append style={draw=#1}
    },
    connect zoomboxes/.style={
    spy connection path={\draw[draw=none,zoombox paths] (tikzspyonnode) -- (tikzspyinnode);}
    },
    help grid code/.code={
        \begin{scope}[
                x={(image.south east)},
                y={(image.north west)},
                font=\footnotesize,
                help lines,
                overlay
            ]
            \foreach \x in {0,1,...,9} { 
                \draw(\x/10,0) -- (\x/10,1);
                \node [anchor=north] at (\x/10,0) {0.\x};
            }
            \foreach \y in {0,1,...,9} {
                \draw(0,\y/10) -- (1,\y/10);                        \node [anchor=east] at (0,\y/10) {0.\y};
            }
        \end{scope}    
    },
    help grid/.style={
        append after command={
            [help grid code]
        }
    },
}

\newcommand\phantomimage{%
    \phantom{%
        \rule{\imagewidth}{\imageheight}%
    }%
}
\newcommand\zoombox[2][]{
    \begin{scope}[zoombox paths]
        \pgfmathsetmacro\xpos{
            (\columncount-1)*(\imagewidth / \pgfkeysvalueof{/tikz/zoomboxarray columns} + \pgfkeysvalueof{/tikz/zoomboxarray inner gap} / \pgfkeysvalueof{/tikz/zoomboxarray columns} ) + \pgflinewidth
        }
        \pgfmathsetmacro\ypos{
            (\rowcount-1)*( \imageheight / \pgfkeysvalueof{/tikz/zoomboxarray rows} + \pgfkeysvalueof{/tikz/zoomboxarray inner gap} / \pgfkeysvalueof{/tikz/zoomboxarray rows} ) + 0.5*\pgflinewidth
        }
        \edef\dospy{\noexpand\spy [
            #1,
            zoombox paths/.append style={
                black and white pattern=\patternnumber
            },
            every spy on node/.append style={#1},
            x=\imagewidth,
            y=\imageheight
        ] on (#2) in node [anchor=north west] at ($(zoomboxes container.north west)+(\xpos pt,-\ypos pt)$);}
        \dospy
        \pgfmathtruncatemacro\pgfmathresult{ifthenelse(\columncount==\pgfkeysvalueof{/tikz/zoomboxarray columns},\rowcount+1,\rowcount)}
        \global\let\rowcount=\pgfmathresult
        \pgfmathtruncatemacro\pgfmathresult{ifthenelse(\columncount==\pgfkeysvalueof{/tikz/zoomboxarray columns},1,\columncount+1)}
        \global\let\columncount=\pgfmathresult
        \ifblackandwhitecycle
            \pgfmathtruncatemacro{\newpatternnumber}{\patternnumber+1}
            \global\edef\patternnumber{\newpatternnumber}
        \fi
    \end{scope}
}

\newcommand\blfootnote[1]{%
  \begingroup
  \renewcommand\thefootnote{}\footnote{#1}%
  \addtocounter{footnote}{-1}%
  \endgroup
}

\newcommand{\whitebox}{\hfill\textcolor{white}{\rule[1.2mm]{1.2mm}{1.2mm}}\hfill}
\newcommand{\filmbox}[1]{%
    \setlength{\fboxsep}{0pt}%
    \colorbox{black}{%
        \begin{minipage}{3.2cm}
            \rule{0mm}{3.2mm}\whitebox\whitebox\whitebox\whitebox\whitebox%
            \whitebox\whitebox\whitebox\whitebox\null\\%
            \null\hfill\includegraphics[width=3.15cm]{#1}\hfill\null\\[0mm]%
            \null\whitebox\whitebox\whitebox\whitebox\whitebox%
            \whitebox\whitebox\whitebox\whitebox\null
        \end{minipage}}}

\definecolor{cvprblue}{rgb}{0.21,0.49,0.74}
\usepackage[pagebackref,breaklinks,colorlinks,citecolor=cvprblue]{hyperref}

\def\paperID{10410} 
\def\confName{CVPR}
\def\confYear{2024}

\title{Selective, Interpretable and Motion Consistent \\Privacy Attribute Obfuscation for Action Recognition}

\author{Filip Ilic\\
TU Graz\\
{\tt\small filip.ilic@tugraz.at}
\and
He Zhao\\
York University\\
{\tt\small zhufl@eecs.yorku.ca}
\and
Thomas Pock\\
TU Graz\\
{\tt\small pock@tugraz.at}
\and
Richard P. Wildes\\
York University\\
{\tt\small wildes@cse.york.ca}
}

\begin{document}

\twocolumn[{%
\renewcommand\twocolumn[1][]{#1}%
\maketitle
\vspace{-0.7cm}
\begin{center}
    \centering
    \captionsetup{type=figure}
    \resizebox{1.0\linewidth}{!}{
        \begin{tikzpicture}[spy using outlines={circle, ultra thick, green!40 ,magnification=4.4,size=2.5cm, connect spies}]
            \setlength{\fboxsep}{0pt}
            \begin{scope}
                \node[inner sep=0pt] (one) at (0,-1) {\fbox{\includegraphics[width=.25\textwidth]{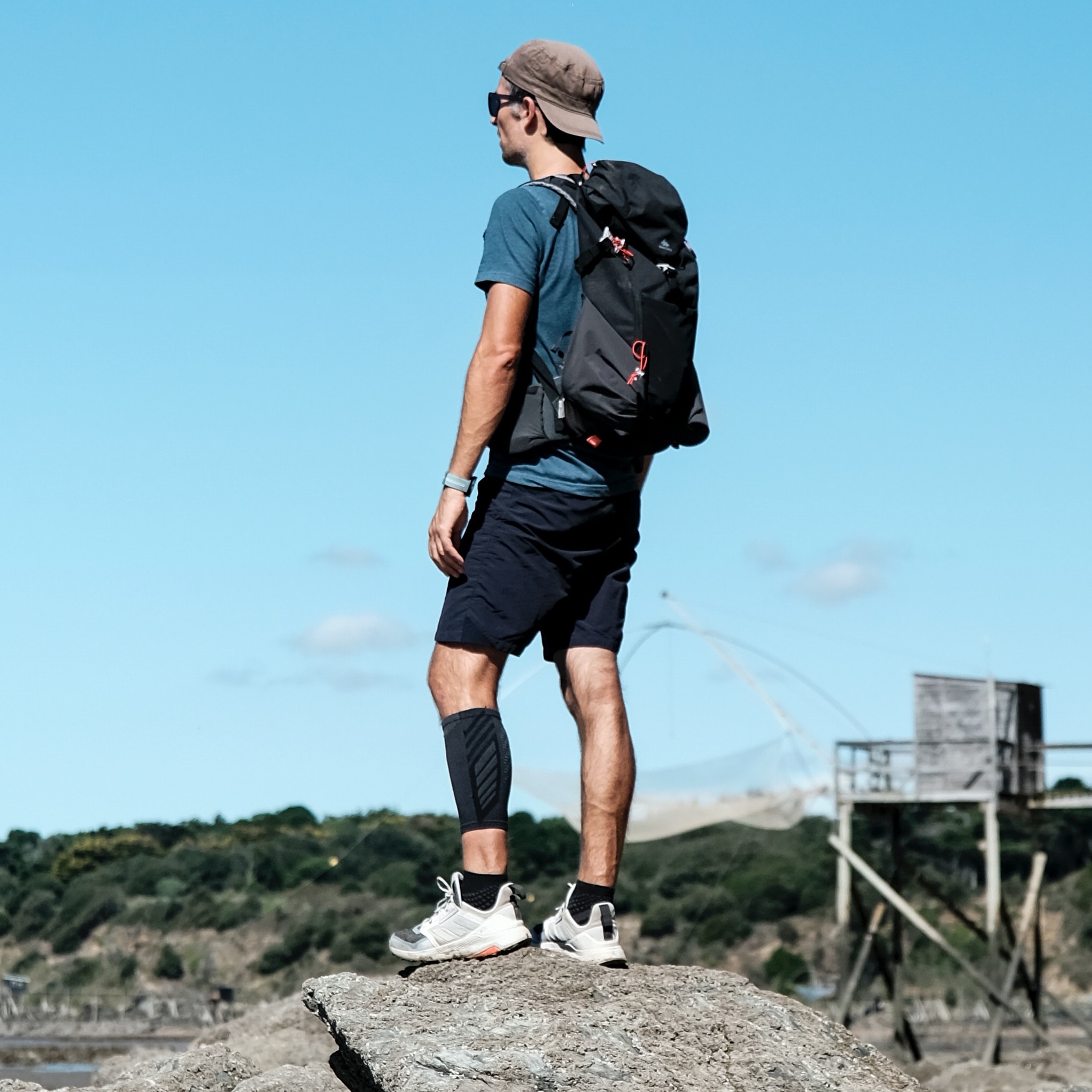}}};
                \node[inner sep=0pt] (three) at (-1.5, 1.0) {\fbox{\includegraphics[width=.25\textwidth]{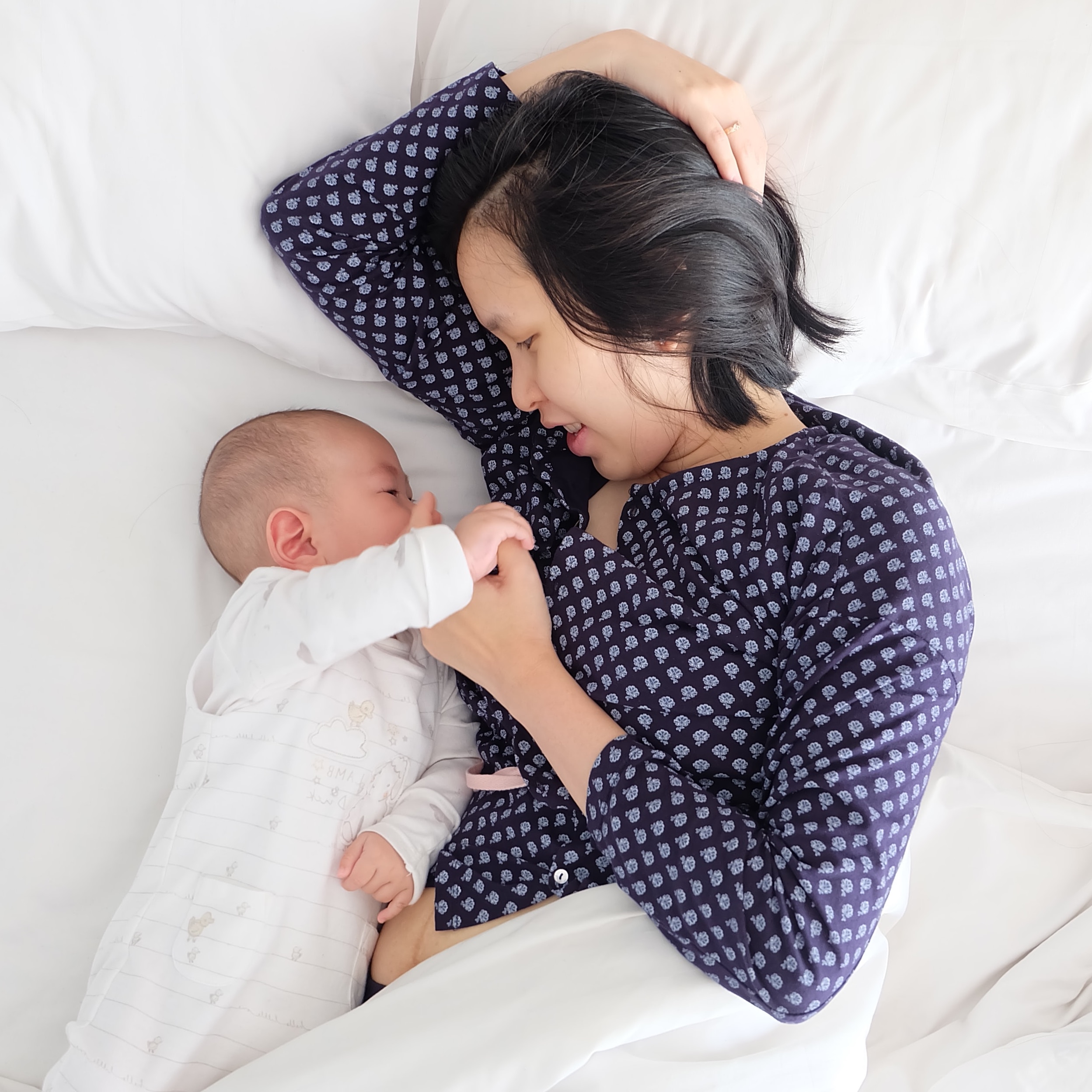}}};
                \node[inner sep=0pt] (two) at (2.5, 1.5) {\fbox{\includegraphics[width=.25\textwidth]{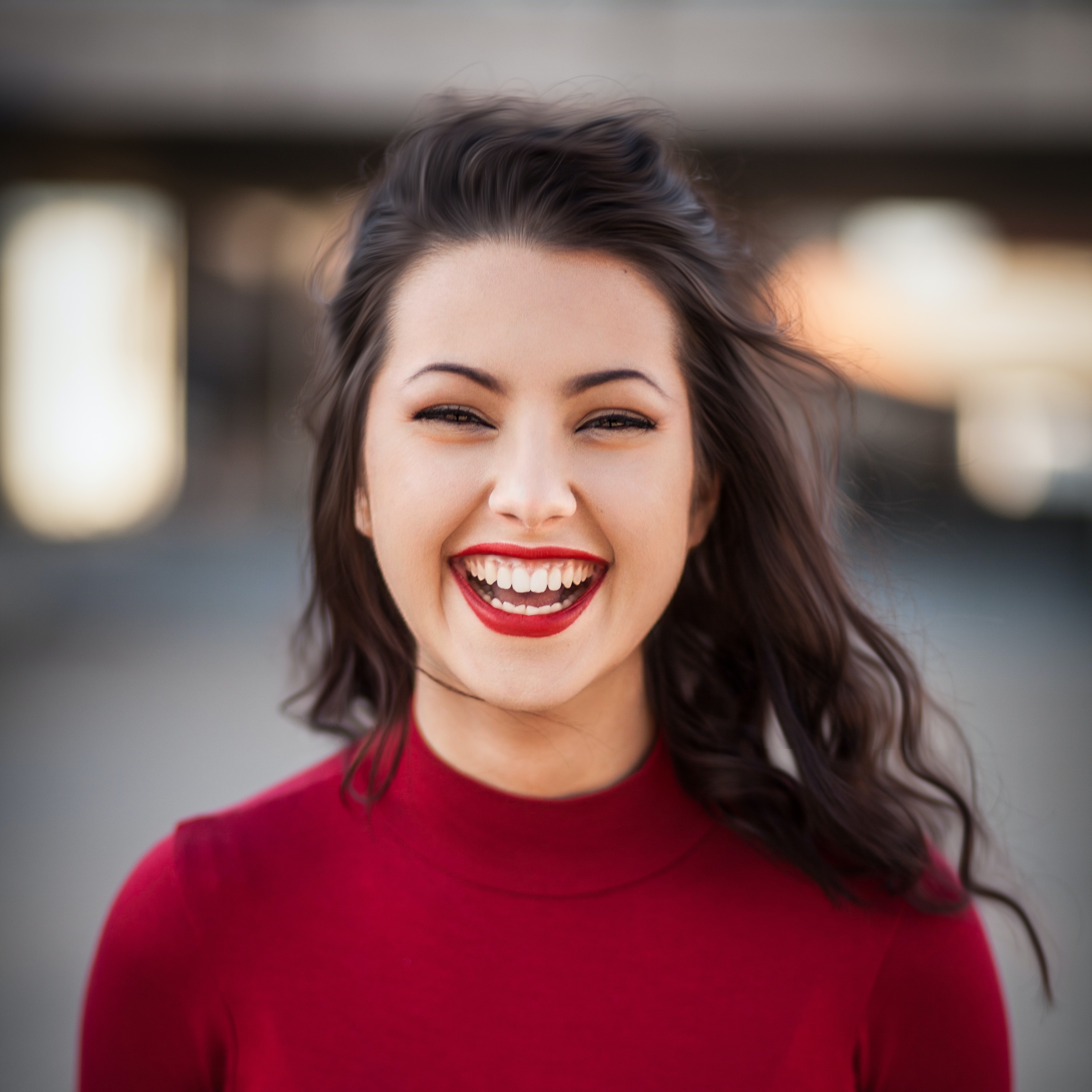}}};
            \end{scope}

            \node[overlay] at (0.5,4.2) {\LARGE{Arbitrary Templates}};
            \node[overlay] at (12.0,4.2) {\LARGE{Source}};
            \node[overlay] at (12.0,-0.8) {\LARGE{Saliency Map}};
            \node[overlay] at (24,4.2) {\LARGE{Obfuscation}};
            \node[overlay] at (24,-0.8) {\LARGE{Optical Flow}};
            \node[overlay] at (24, 4.9) {\LARGE{Temporally Consistent}};

            \draw[thick] (5.9,-5) -- (5.9,4);

            \node[inner sep=0pt] (tmp) at (12,-3) {\filmbox{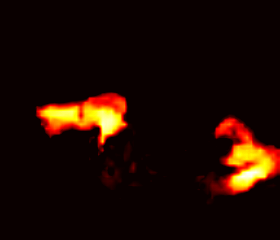}\filmbox{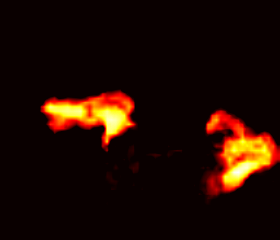}\filmbox{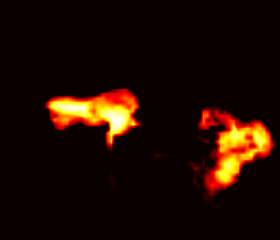}};
            \node[inner sep=0pt] (tmp) at (12,2) {\filmbox{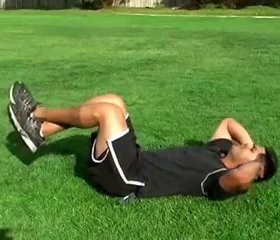}\filmbox{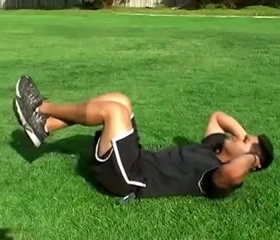}\filmbox{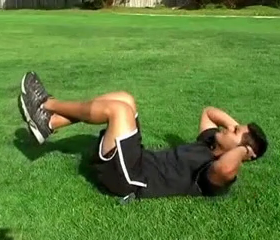}};
            \node[inner sep=0pt] (tmp) at (24,2) {\filmbox{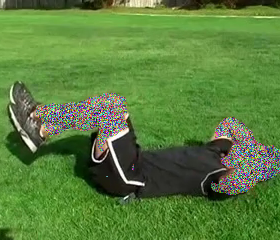}\filmbox{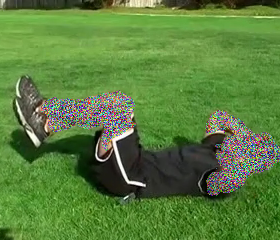}\filmbox{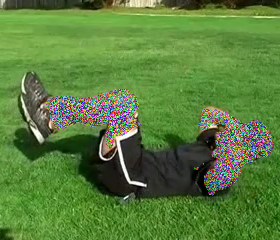}};   
            \node[inner sep=0pt] (tmp) at (24,-3) {\filmbox{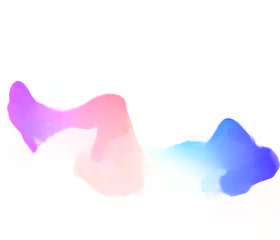}\filmbox{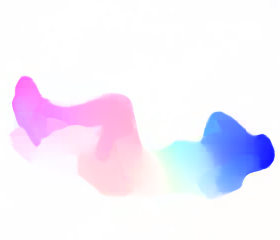}\filmbox{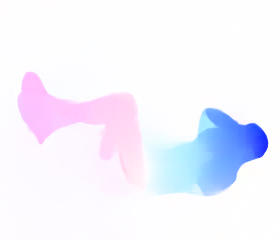}};   

            \spy on (-2.4, 1.1) in node at (-3.2,-4); 
            \spy on (-0.27, -1.5) in node at (-1,-4); 
            \spy on (2.45, 1.75) in node at (1.2,-4); 
            \spy on (2.8, 2) in node at (3.4,-4); 

            \spy [green!40, ultra thick, circle, magnification=4, size=4.0cm] on (13.05, 1.75) in node at (15.8, 3.);
            \spy [green!40, ultra thick, circle, magnification=4, size=4.0cm] on (13.05, -3.15) in node at (15.8, -2.);
            \spy [green!40, ultra thick, circle, magnification=4, size=4.0cm] on (25.05, 1.75) in node at (27.8, 3.);
            \spy [green!40, ultra thick, circle, magnification=4, size=4.0cm] on (25.05, -3.15) in node at (27.8, -2.);
        \end{tikzpicture}
    }
    \captionof{figure}{Our goal is to hide privacy attributes without action recognition performance dropping. 
    Left: Arbitrary images can be used to specify an interpretable template library defined by privacy attributes. Middle: A salience map is generated from privacy templates; example illustrates use of templates for personal identification. Right: The source video is masked with noise as guided by salience and animated by source video optical flow. Salience makes masking selective to privacy sensitive regions, while preserving \textit{scene context}; optical flow preserves \textit{motion} -- both of which are critical for action recognition.  The obfuscated video can be input directly to arbitrary privacy and action recognition systems without retraining. Zoomed circles highlight details only for illustration.
    }
    \label{fig:teaser}
\end{center}%
}]
\begin{abstract}
\vspace{-1em}
Concerns for the privacy of individuals captured in public imagery have led to privacy-preserving action recognition. Existing approaches often suffer from issues arising through obfuscation being applied globally and a lack of interpretability. Global obfuscation hides privacy sensitive regions, but also contextual regions
important for action recognition. Lack of interpretability erodes trust in these new technologies. We highlight the limitations of current paradigms and propose a solution: Human selected privacy templates that yield \textit{interpretability by design}, an obfuscation scheme that \textit{selectively hides attributes} and also induces \textit{temporal consistency}, which is important in action recognition. Our approach is architecture agnostic and directly modifies input imagery, while existing approaches generally require architecture training. Our approach offers more flexibility, as no training is required, and outperforms alternatives on three widely used datasets.%
\blfootnote{Code available \href{https://f-ilic.github.io/SelectivePrivacyPreservation}{f-ilic.github.io/SelectivePrivacyPreservation}}
\end{abstract}
\section{Introduction}\label{sec:into}
\label{subsec:motivation}
Advances in state-of-the-art computer vision and machine learning enable deployment of such systems in the public sphere. Accompanying these initiatives, concerns arise for the  privacy of individuals that are captured in acquired imagery~\cite{imgdatasetsocial, scalableoffensive, genderimb, safediff}.
In particular, considerations arise regarding attributes that individuals want to keep confidential, yet that are revealed through visual information, even though they are not critical for the functioning of the deployed system, \eg identity, age, gender and race.
Video-based action recognition is an area of consideration as it has potential for widespread applications in surveillance and monitoring.
These concerns have sparked interest in privacy preserving action recognition \cite{prilens, priaction, spact, bqn}. These approaches process input imagery to obscure privacy attributes while maintaining action recognition performance. Contemporary approaches typically apply their obfuscation across entire input video frames and improvements have been made within this paradigm.
Notably, however, there are downsides to this paradigm, as follows.

\vspace{0.2em}
\noindent\textbf{Collateral damage.} Global masking strategies indiscriminately obscure the entire image, impacting regions within the scene that may exhibit high correlations with actions, albeit lack relevance to privacy. For example, it is known that masking objects and scene context can impair action recognition performance~\cite{contextaction}, yet these are lost in global masking. Furthermore,  global masking strategies do not allow for selective attribute obfuscation and generally hide all attributes at once, even when not all are of concern.

\vspace{0.2em}
\noindent\textbf{Loss of dynamic information.} The large change in input modalities from global masking necessitates the retraining of the action recognition module or the design of custom modules (adversarial training), which adds to the challenge of practical deployment. Indeed, even if applied more locally, the obfuscation can compromise the motion of the actors, which also can be important in action recognition~\cite{simonyan2014two}. 

\vspace{0.2em}
\noindent\textbf{Lack of interpretability.} Finally, given that state-of-the-art approaches are end-to-end trained with limited concern for interpretability, the exact nature of what is being masked and how it is achieved can remain unclear. Lack of interpretability is an important concern in privacy preservation, because its lack can compromise user trust~\cite{karran2022designing, liu2022trustworthy}.

\subsection{Contributions}\label{subsec:contributions}
We present 
an approach to privacy preserving action recognition that responds directly to current limitations in four ways, as illustrated in Fig.~\ref{fig:teaser}. (i) 
The approach is based on local detection and selective obfuscation of privacy sensitive regions. This selectivity maintains global context information that is crucial to action recognition, yet unimportant for privacy. 
(ii) The local processing avoids large modality shifts in the imagery and is independent of the action recognition module itself; therefore, it does not require  algorithm retraining, which is sometimes infeasible. (iii) The masking preserves interframe motion; so, that information is available for action recognition. (iv) 
The privacy sensitive masking is interpretable by design, and allows inspection through the explicitly generated saliency maps.

\section{Related work}
\label{sec:related}
\noindent\textbf{Privacy in machine learning.} 
The need to protect privacy has garnered increased attention in the vision research community. Current models commonly consume large amounts of web data 
to learn generalizable representations~\cite{imagenet, yfcc100m, ego4d}, which inevitably invade personal information, such as identity and location. Moreover, in model deployment it also may be desirable to preserve privacy information. The concern for privacy is not limited to vision research, but extends across artificial intelligence, including natural language processing~\cite{dpLLMs} and more general machine learning~\cite{dpML}. As these technologies find their way into broader society, privacy concerns must be considered.

\noindent\textbf{Privacy preserving action recognition.} 
We focus on video-based action recognition. Recent developments in this area have yielded systems capable of strong performance on challenging datasets; for review see, \eg \cite{sharma22}. 
Similarly, applications, including those in privacy-sensitive scenarios (\eg surveillance~\cite{videosurveillance} and monitoring~\cite{anomaly}), are being developed. 
To provide useful spatiotemporal signals for action recognition, video clips are mostly collected to capture actors with a great level of clarity throughout the actions, thus increasing the chance of privacy leakage. Moreover, such datasets are expanding rapidly, \eg~\cite{Kinetics, ActivityNet}. 

In response to these concerns, research on recognizing actions while preserving private information has emerged in recent years. Early work concentrated on devising models that can work on low-resolution videos~\cite{ryoo2017privacy, ryoo2018extremelowres, chen2017semi}; these methods often operate at the cost of sacrificing action recognition performance. Other work along these lines developed face-anonymization techniques to prevent models from yielding high accuracy on face recognition~\cite{ren2018anonmface, jourabloo2015anonmattribute}. Still, those approaches cannot easily extend to other attributes and are correspondingly limited. More recent work has focused on implicitly learning transformation functions to anonymize videos in a data-driven fashion~\cite{pahmdb, spact, prilens, iccv23privacy}, and also extending such approaches to anomaly detection~\cite{iccv23privacyanomaly}. To reduce model complexity, a competing approach follows a simple, yet effective procedure~\cite{bqn}: Frame subtraction, followed by broadband-filtering to yield motion descriptors for action recognition, while suppressing privacy attributes. A notable limitation of that approach is the relative weakness of frame differences as motion descriptors compared to common alternatives, \eg optical flow. Another common limitation to all existing work is the lack of flexibility to select arbitrary private information to hide, \ie preserving privacy of only critical attributes, while maintaining the visibility of others, to avoid obfuscation of visual cues that are essential to action classification.

\noindent\textbf{Template matching.} 
Building correspondences between templates and target images via matching is a foundation for modern vision research, \eg SfM~\cite{multipleviewgeo}, image correspondences~\cite{superglue}, detection~\cite{hogdetection} and tracking~\cite{goodfeature}. Features used in matching have advanced rapidly from primarily hand-crafted~ (\eg \cite{hogdetection, sift}) to learning-based convolutional \cite{resnet} or transformers \cite{dosovitskiy2020image}. Our solution is inspired by techniques seen in feature-based template matching. We apply their insights on matching templates to localize and selectively obfuscate privacy-sensitive regions. More specifically, we adopt DINO-ViT features~\cite{caron2021emerging} and compute similarities using the \textit{keys} of the last attention layer from that architecture. Our use of DINO-ViT keys (rather than queries or values) is based on previous work finding them to perform best when applied to visual correspondence~\cite{amir2021deep, oquab2023dinov2}. 
\begin{figure*}[h]

	\newlength\figHeight%
    \setlength\figHeight{3.0cm}%
	\centering
	\subcaptionbox{Overview\label{fig:overview}}{%
       \includegraphics[height=3.0cm]{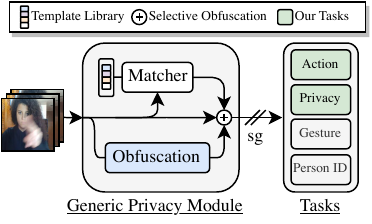}
	}%
	\hfill%
	\tikz{\draw(0,\figHeight) -- (0,0);}%
	\hfill%
	\subcaptionbox{Matcher\label{fig:matching}}{%
        \includegraphics[height=3.0cm]{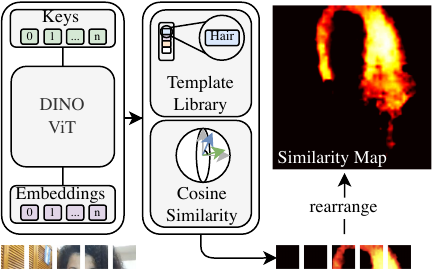}
	}%
	\hfill%
	\tikz{\draw(0,\figHeight) -- (0,0);}%
	\hfill%
		\subcaptionbox{Temporally Consistent Selective Obfuscation\label{fig:selective}}{%
		\includegraphics[height=3.0cm]{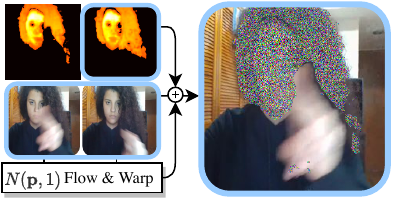}
	}%
    \caption{Overview of Method. \ref{fig:overview} We present a privacy module that builds atop three components: (i) a semantic \textit{template library} that contains attributes to be hidden, (ii) a descriptor \textit{matcher} to localize template features in videos to be obscured and, (iii) an \textit{obfuscation} method that is sensitive with respect to motion present in the scene. 
    \ref{fig:matching} A semantic descriptor matcher based on DINO~\cite{caron2021emerging}-ViT~\cite{dosovitskiy2020image} keys is used to determine privacy salient regions in a video based on the template library. In our case, regions of interest correspond to those that can identify a person; however, this component can be adapted for other privacy attributes through specification of different templates. The result is a saliency map. 
    \ref{fig:selective}
    The saliency map is  used as a weight to apply noise to the regions.
    The noise, however, is not static, but is warped with optical flow with an initial noise pattern image, $N(\mathbf{p}, 1)$, for the purposes of preserving motion information in the source video. The similarity maps of all aggregated relevant privacy attributes are used to weigh the noise and apply it to the input image, obfuscating privacy sensitive information while not destroying the underlying temporal signal.}
   \label{fig:method}
\end{figure*}

\section{Technical approach}
\label{sec:method}

Our approach to privacy preserving action recognition is a stand alone module that operates by preprocessing video that subsequently is input to  arbitrary action and privacy recognition algorithms, \ie it is independent of the recognition algorithms and does not entail any retraining of those algorithms. Our method consists of three key components; see Fig.~\ref{fig:overview}: (i) A template library that covers privacy attributes to be preserved, 
(ii) a matcher that produces saliency maps 
between selected templates and images 
where privacy is to be preserved and 
(iii) an obfuscator that uses the saliency maps to hide privacy attributes of concern in a temporally consistent fashion to preserve motion in the video. The remainder of this section details each of these components.

\subsection{Template library}
\label{sec:templatelibrary}
\begin{figure}[!t]
	\centering
	\begin{tikzpicture}[spy using outlines={green!40,magnification=4.2,size=0.8cm}]
		\node (face) {\pgfimage[interpolate=true,height=2.8cm]{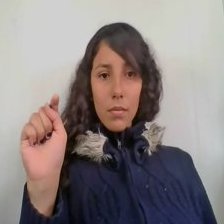}};
		\spy[every spy on node/.append style={ultra thick}]  on (-0.1, 0.75) in node [left] at (-1.6,1.0); 
		\spy[every spy on node/.append style={ultra thick}]  on (-0.1, 0.45) in node [left] at (-1.6,0); 
		\spy[every spy on node/.append style={ultra thick}]  on (-0.85, -0.23) in node [left] at (-1.6,-1.0); 
		\spy[every spy on node/.append style={ultra thick}]  on (0.07, 0.01) in node [left] at (-2.6,-1.0); 
		\spy[every spy on node/.append style={ultra thick}]  on (0.35, 0.7) in node [left] at (-2.6,1.0); 
		\spy[every spy on node/.append style={ultra thick}]  on (0.3, 0.25) in node [left] at (-2.6,0);
	\end{tikzpicture}%
	\hfill
	\begin{tikzpicture}[spy using outlines={green!40,magnification=4.2,size=0.8cm}]
		\node (face) {\pgfimage[interpolate=true,height=2.8cm]{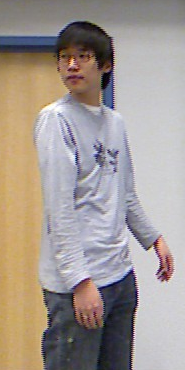}};
		\spy[every spy on node/.append style={ultra thick}]  on (-0.2,-1.2) in node [left] at (-0.9,-1.0);
		\spy[every spy on node/.append style={ultra thick}]  on (0.05, 0) in node [left] at (-0.9,1.0);
		\spy[every spy on node/.append style={ultra thick}]  on (-0.21, -0.35) in node [left] at (-0.9,0);
	\end{tikzpicture}%
	\caption{Template Library consisting of Patches Chosen from Anatomical Landmark Regions. These specific images are passed through a DINO-ViT feature extractor. The keys, corresponding to spatial locations of the highlighted patches, are chosen as the templates for matching to input images to obtain semantically similar regions for obfuscation.
}
	\label{fig:templates}
\end{figure}

To obfuscate privacy sensitive regions in images in an interpretable fashion, we need an explicit set of ``attributes" to be hidden from the target video. Such a concept template library, $\mathsf{T}$, can be built manually choosing image patches corresponding to features one wishes to obfuscate. In this paper, we concentrate on privacy attributes related to personal identification. Therfore, we use landmark anatomical features, corresponding to detailed facial landmarks and the hand (Fig.~\ref{fig:templates} left: \textit{forehead, hair, eye, cheek, lips, hand}) as well as larger body parts (Fig.~\ref{fig:templates} right: \textit{arm, torso, legs}).
The choice to focus on preservation of human identity is motivated by the fact that all three of our evaluation datasets define their privacy attributes on attributes relating to person identity. Notably, however,
such a template library easily can be extended depending on the particular task at hand. For example, attributes pertaining to location (e.g. street signs, distinctive scene objects) could guide development of a complimentary set of templates. In any case, given a template library, a user can selectively combine templates to obscure attributes of concern in a particular application. 

Formally, we define each template $\tau_i$ as an element in the set $\mathsf{T} = \{\tau_1, \tau_2, ..., \tau_n\}$. Further, let $\tilde{\mathsf{T}} \subseteq \mathsf{T}$ be the subset of templates that the user selects from the library, $\mathsf{T}$, for a particular subset of privacy attributes to be preserved for a given dataset or application.
Our manual approach to template selection leads to concepts that are \textit{interpretable by design}. In particular, we use two source images, shown in Fig.~\ref{fig:templates}, one for small-scale features on the face, and one for large regions, taken from the IPN \cite{ipn} and SBU \cite{sbu} datasets. We choose these two datasets because of their complimentarity in template selection: IPN focuses on small scale features (\eg facial and hand), while SBU focuses on larger scale features (\eg larger body parts), as detailed later in Sec.~\ref{sec:protocol}. 
The choice of the particular template images is not critical to the functioning of our approach, because we extract semantic features from the templates that are known to generalize well across images instances~\cite{caron2021emerging}, as described next. 

\subsection{Matching: Local patch descriptor templates}\label{sec:matching}
\noindent\textbf{Semantic features.} 
Our requirements for good features to match between privacy templates in our library and frames in an action recognition video are straightforward: We require (i) semantically rich features that generalize across (in our case) different individuals and (ii) high spatial resolution to perform privacy obfuscation in a localized manner without destroying regions that are required for action recognition. We find that DINO-ViT features fit our needs very well~\cite{caron2021emerging,vaswani2017attention}: (i) They have been trained to yield high similarity for semantically related concepts (\eg objects and their parts), while suppressing the similarity of unrelated matters (\eg background). (ii) They support local patch feature extraction over high resolution images without losing global context.
Elsewhere, DINO-ViT keys have proven especially useful in matching between templates and target images~\cite{amir2021deep, oquab2023dinov2}. 
Following these advances, we use last attention layer DINO-ViT keys computed from privacy templates (\eg Fig.~\ref{fig:templates}) to match against input images to compute privacy saliency maps.

\begin{figure*}
    \centering
    \includegraphics[width=\linewidth]{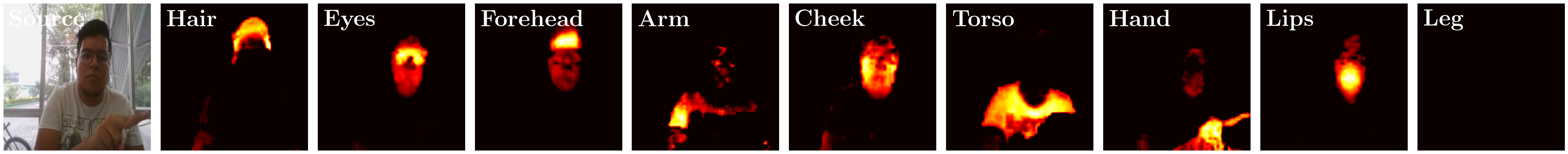}\\
    \includegraphics[width=\linewidth]{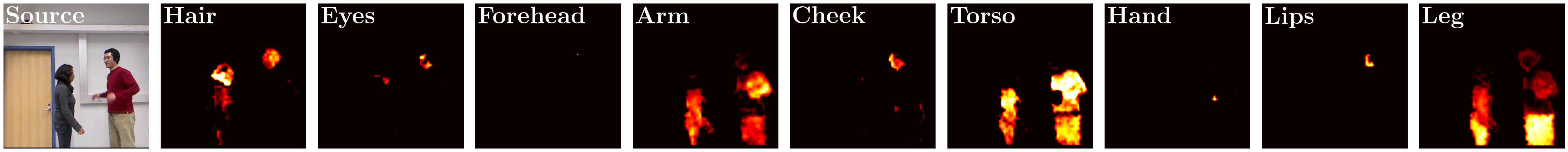}\\
    \vspace{-0.5em}
    \caption{Saliency Maps for Descriptors in the Template Library. The manual selection of these templates allows for \textit{interpretability} of the obfuscated parts of the image \textit{by design}. The matched DINO-ViT features capture rich semantic information and allow for detailed spatial localization due to the nature of vision transformers. These saliency maps can then be combined for obfuscating any combination of templates depending on the task at hand.}
    \label{fig:saliency}
\end{figure*}

\vspace{0.5em}
\noindent\textbf{Privacy saliency matching.}
To find privacy salient regions in an input video, a similarity map, $S$, is computed between each frame, $I$, in the video and the selected set of privacy templates, $\tilde{\mathsf{T}}$. The image is tiled with $m$ patches, $I_j, j \in \{1,...,m\}$, and for each patch DINO-ViT keys, $\sf{K}$$(I_j)$, are extracted.
These features are matched with DINO-ViT keys, $\sf{K}$$(\tau_i)$, for all selected templates, $\tau_i \in\tilde{\mathsf{T}}$. Clipped cosine similarity  is used to establish the saliency of each image patch according to
\begin{equation}
\text{\small
    $s_j = \frac{1}{|\tilde{\mathsf{T}}|} \sum_{i=1}^{|\tilde{\mathsf{T}}|} \max\left(0, \frac{ \langle \mathsf{K}(\tau_i), \mathsf{K}(I_j)\rangle}{\|\mathsf{K}(\tau_i)\|\|\mathsf{K}(I_j)\|}\right), \forall j \in \llbracket
1,m\rrbracket
,$}
\label{eq:saliency}
\end{equation}
where $\langle \cdot , \cdot \rangle$ is inner product and $|\tilde{\mathsf{T}}|$ is the cardinality of $\tilde{\mathsf{T}}$, the set of selected privacy templates.
Clipping is used because only positive values imply salience.
This calculation is performed for every patch in every image of the video. Subsequently, the $m$ saliency patches are reassembled in the shape of the original image to produce the final saliency map, 
\begin{equation} \label{eq:saliencyMap}
S = \mathcal{R}(s_1,...,s_m;h,w),    
\end{equation}
with $\mathcal{R}$ a function that accepts image tiles, $s_j$, and reshapes them into their original image format of height $h$ and width $w$. Note that a separate salience map, $S$, is calculated for each frame in a video of interest, $I$. The entire process of feature extraction, matching and the resulting saliency maps are summarized in Fig.~\ref{fig:matching}. Example saliency maps in Fig.~\ref{fig:saliency} illustrate the ability of our approach to capture all of our templates in a variety of scenarios.

\subsection{Temporally consistent obfuscation}\label{sec:obfuscation}
Our goal is to mask privacy sensitive regions in images to obfuscate them.
If we were to apply masks to every frame independently, then temporal information important to action recognition, \eg motion of actors, would be destroyed as well. We empirically document the issue in Sec.~\ref{sec:motionnoise}. In response to this challenge, we follow previous work that presented a method for producing temporally consistent spatial noise across videos that supported action recognition, while obscuring appearance information in single frames~\cite{ilic2022appearance}. While the previous work applied noise patterns uniformly across entire frames, we instead weight them by our privacy salience maps, $S$, to preserve as much context information as possible. The remainder of this subsection details our approach, with an outline shown in Fig.~\ref{fig:selective}.

\vspace{0.5em}
\noindent\textbf{Noise pattern initialization.} Let $\mathbf{p} = (x,y)$ be image coordinates and $t$ time. We initialize a noise image, $N(\mathbf{p}, 1)$, with the same dimensions as a frame from the input video (\ie $h \times w \times 3$), with $h$ height, $w$ width and 3 the number of colour channels. The dataset mean, $\mu$, and standard deviation, $\sigma$, are used to define a uniform distribution from which individual pixel intensities are drawn according to
\begin{equation}\label{eq:afdinit}
N(\mathbf{p},t=1) \sim \mathcal{U}[\mu-\sigma, \mu+\sigma].
\end{equation}

\vspace{0.5em}
\noindent\textbf{Motion consistent noise.} To create \textit{motion consistent noise} the initial random frame, $N(\mathbf{p}, 1)$, is warped forward with flow fields derived from the original video, $I(\mathbf{p},t)$, as extracted by an optical flow algorithm. Let $\mathbf{v}(\mathbf{p},t)=\left(u(\mathbf{p},t),v(\mathbf{p},t)\right)$ be the flow field that maps points, $\mathbf{p}$, in frame $t$ to those in frame $t-1$, with $u$ and $v$ the horizontal and vertical components of the flow. Then, a motion consistent noise sequence is generated as
\begin{equation}\label{eq:rwarp}
	N(\mathbf{p},t)= N\big(\mathbf{p} + \mathbf{v}(\mathbf{p}, t), 1\big)
\end{equation}
with $t \in \{2,...,T\}$ and $T$ the number of frames in the original input video.

\vspace{0.5em}
\noindent\textbf{Selective privacy obfuscation.} The computed video sequence, $N$,
shows no single frame appearance related to the original video as it is random noise; however, when viewed as a video it reveals the motion present in the original, \cf~\cite{ilic2022appearance}. Direct use of this synthesized video obscures privacy attributes; however, it also obscures other context information that could be of use in action recognition. So, instead we selectively apply $N$ to every frame $I$ in the video by using the privacy salience maps $S$, according to
\begin{equation}  \label{eq:rselective}
     O(\mathbf{p},t) = I(\mathbf{p},t) + \Big(S(\mathbf{p},t) \times \big(N(\mathbf{p},t) - I(\mathbf{p},t)\big)\Big).
\end{equation}
The resulting video, $O$, contains selectively obfuscated regions, built with interpretable templates by design and contains motion information that (in principle) does not differ from the original input video.
\section{Empirical evaluation}
\label{sec:evaluation}

The privacy obfuscated video, \eqref{eq:rselective}, serves directly as input to action and privacy recognition. No specialized development, training or other modification of the recognition algorithms nor adaptation of the privacy preservation system is necessary. Indeed, a major difference compared to competing state-of-the-art approaches (\eg~\cite{ryoo2017privacy,pahmdb,bqn}) is that \textit{we do not retrain} networks with our obfuscated data, while \textit{they do retrain}. 
We exploit the fact that privacy attributes are generally \textit{independent} from action recognition cues, which is enabled by our unique \textit{selective obfuscation} approach. 

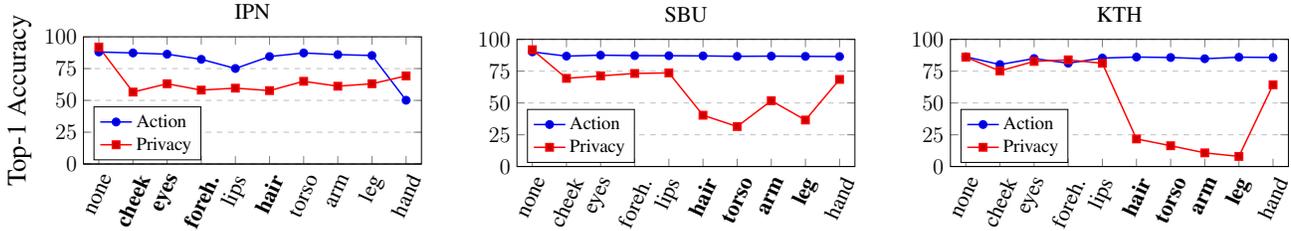
\begin{figure*}
	\centering
	\resizebox{0.33\linewidth}{!}{%
	\begin{tikzpicture}
		\begin{axis}[
			title=\large{IPN},
			ylabel={Top-1 Accuracy},
			bar width=4pt,
			height = 4cm,
			width = 8cm,
			symbolic x coords={
				none,
				cheek,
				eyes,
				foreh.,
				lips,
				hair,
				torso,
				arm,
				leg,
				hand,
			},
			xticklabels={
				none,
				\textbf{cheek},
				\textbf{eyes},
				\textbf{foreh.},
				lips,
				\textbf{hair},
				torso,
				arm,
				leg,	
				hand,
			}, 
			enlarge x limits=0.05,
			xtick = data,
			ytick={0,25,50,75,100},
			ymin=0, ymax=100,
			ymajorgrids=true,
			grid style=dashed,
			legend pos=south west,
			legend cell align={left},
			x tick label style={rotate=65},
			every axis plot/.append style={mark=square*,thick},
			tick label style={font=\large},
			label style={font=\Large}
			]

		\draw (axis cs:{[normalized]\pgfkeysvalueof{/pgfplots/xmin}},0)
            -- (axis cs:{[normalized]\pgfkeysvalueof{/pgfplots/xmax}},0);

		\addplot+[]
		coordinates {
			(none,88.05)
			(cheek,87.28)
			(eyes,86.31)
			(foreh.,82.29)
			(lips,75.07)
			(hair,84.47)
			(torso,87.27)
			(arm,85.97)
			(leg,85.28)
			(hand,50.12)
		};
		\addlegendentry{Action}

		\addplot+[]
		coordinates {
			(none,91.92)
			(cheek,56.54)
			(eyes,63.08)
			(foreh.,58.08)
			(lips,59.62)
			(hair,57.69)
			(torso,65.00)
			(arm,61.15)
			(leg,63.08)
			(hand,69.23)
		};
		\addlegendentry{Privacy}
		\end{axis}
	\end{tikzpicture}
	}%
	\resizebox{0.33\linewidth}{!}{%
	\begin{tikzpicture}
		\begin{axis}[
			title=\large{SBU},
			ylabel={\phantom{Top-1 Accuracy}},
			height = 4cm,
			width = 8cm,
			bar width=4pt,
			symbolic x coords={
				none,
				cheek,
				eyes,
				foreh.,
				lips,
				hair,
				torso,
				arm,
				leg,
				hand,
			},
			xticklabels={
				none,
				cheek,
				eyes,
				foreh.,
				lips,
				\textbf{hair},
				\textbf{torso},
				\textbf{arm},
				\textbf{leg},	
				hand,
			}, 
			enlarge x limits=0.05,
			xtick = data,
			ytick={0,25,50,75,100},
			ymin=0, ymax=100,
			ymajorgrids=true,
			grid style=dashed,
			legend pos=south west,
			legend cell align={left},
			x tick label style={rotate=65},
			every axis plot/.append style={mark=square*,thick},
			tick label style={font=\large},
			label style={font=\Large}
			]

		\draw (axis cs:{[normalized]\pgfkeysvalueof{/pgfplots/xmin}},0)
            -- (axis cs:{[normalized]\pgfkeysvalueof{/pgfplots/xmax}},0);

		\addplot+[]
		coordinates {
			(none,90.15)
			(cheek,86.82)
			(eyes,87.56)
			(foreh.,87.23)
			(lips,87.15)
			(hair,86.99)
			(torso,86.66)
			(arm,86.82)
			(leg,86.66)
			(hand,86.50)
		};
		\addlegendentry{Action}

		\addplot+[]
		coordinates {
			(none,91.87)
			(cheek,69.36)
			(eyes,71.28)
			(foreh.,73.21)
			(lips,73.62)
			(hair,40.36)
			(torso,31.49)
			(arm,51.77)
			(leg,36.60)
			(hand,68.51)
		};
		\addlegendentry{Privacy}
		\end{axis}
	\end{tikzpicture}
	}%
	\resizebox{0.33\linewidth}{!}{%
	\begin{tikzpicture}
		\begin{axis}[
			title=\large{KTH},
			ylabel={\phantom{Top-1 Accuracy}},
			bar width=4pt,
			height = 4cm,
			width = 8cm,
			symbolic x coords={
				none,
				cheek,
				eyes,
				foreh.,
				lips,
				hair,
				torso,
				arm,
				leg,
				hand,
			},
			xticklabels={
				none,
				cheek,
				eyes,
				foreh.,
				lips,
				\textbf{hair},
				\textbf{torso},
				\textbf{arm},
				\textbf{leg},	
				hand,
			}, 
			enlarge x limits=0.05,
			xtick = data,
			ytick={0,25,50,75,100},
			ymin=0, ymax=100,
			ymajorgrids=true,
			grid style=dashed,
			legend pos=south west,
			legend cell align={left},
			x tick label style={rotate=65},
			every axis plot/.append style={mark=square*,thick},
			tick label style={font=\large},
			label style={font=\Large}
			]

		\draw (axis cs:{[normalized]\pgfkeysvalueof{/pgfplots/xmin}},0)
            -- (axis cs:{[normalized]\pgfkeysvalueof{/pgfplots/xmax}},0);

		\addplot+[]
		coordinates {
			(none,86.31)
			(cheek,80.18)
			(eyes,84.92)
			(foreh.,81.28)
			(lips,85.28)
			(hair,86.08)
			(torso,85.72)
			(arm,84.77)
			(leg,85.97)
			(hand,85.77)
		};
		\addlegendentry{Action}

		\addplot+[]
		coordinates {
			(none,85.96)
			(cheek,75.20)
			(eyes,82.67)
			(foreh.,83.87)
			(lips,81.20)
			(hair,21.73)
			(torso,16.40)
			(arm,10.80)
			(leg,8.00)
			(hand,64.27)
		};
		\addlegendentry{Privacy}
		\end{axis}
	\end{tikzpicture}
	}%
        \vspace{-0.5em}
	\caption{Obfuscation with a Single Attribute and the Impact on Performance. Attribute importance is dataset dependent. For example, notice how the 'Hand' template contributes to a large decrease in action recognition performance on IPN, as the action is determined soley by the hand, whereas on SBU it does not. Optimally \textbf{\color{blue} blue} is high and \textbf{\color{red} red} is low. Corresponding qualitative examples of saliency maps for each individual template are shown in Fig.~\ref{fig:saliency}.
 \textbf{Bold} text along the abscissa of each plot  indicates the templates used for the final results.}
	\label{fig:individual}
\end{figure*}
%
%
%
\begin{table}
	\small
	\centering
	\footnotesize{\textsc{Action Recognition on Source Datasets}}\vspace{0.3em}
		\begin{tabularx}{\linewidth}{lYYYY} 
			\toprule
			\textbf{Network} & $f \times r$  & IPN & KTH & SBU\\
			\midrule
			\rc C2D \cite{wang2018non}& $ 8 \times 8 $ & 74.56 & 83.00 & 87.50 \\
			CSN \cite{tran2019video}& $ 32 \times 2 $ & 91.73 & 88.33 & 90.91 \\
			\rc E2S X3D L \cite{ilic2022appearance}& $ 16 \times 5 $ & \textbf{95.15} & \textbf{95.33} &  \underline{91.11} \\
			E2S X3D M \cite{ilic2022appearance}& $ 16 \times 5 $ & 89.38 & 92.33 & 86.36\\
			\rc E2S X3D S \cite{ilic2022appearance}& $ 13 \times 6 $ & 85.16 & 92.00 & 82.98 \\
			I3D \cite{carreira2017quo} & $ 8 \times 8 $ & 83.15 & 89.67 & 82.95 \\
			\rc MVIT\cite{fan2021multiscale} & $ 16 \times 4 $ & 88.00 & 90.00 & \textbf{92.55}\\
			R2+1D \cite{tran2018closer} & $ 16 \times 4 $ & 89.80 & 87.33 & 81.91 \\
			\rc Slow \cite{feichtenhofer2019slowfast}& $ 8 \times 8 $ & 85.76 & 89.00 & 88.64 \\
			SlowFast \cite{feichtenhofer2019slowfast}& $ 32 \times 2 $ & 88.06 & 87.33 & 90.00 \\
			\rc X3D L \cite{feichtenhofer2020x3d}& $ 16 \times 5 $ & \underline{94.41} & \underline{94.00} & 90.22 \\
			X3D M \cite{feichtenhofer2020x3d}& $ 16 \times 5 $ & 91.67 & 93.00 &  81.91 \\
			\rc X3D S \cite{feichtenhofer2020x3d}& $ 13 \times 6 $ & 87.81 & 90.67 & 75.00 \\
			\midrule
			Average&  & 88.05 & 90.15 & 86.31 \\
			\bottomrule
		\end{tabularx}\\
		\vspace{0.3em}\footnotesize{\textsc{Privacy Preservation for Source Datasets}}\vspace{0.3em}
		\begin{tabularx}{\linewidth}{lYYYY} 
			\toprule
			\textbf{Network} & IPN & KTH& SBU\\
			\midrule
			\rc ResNet$_{18}$ \cite{he2016deep}  & 88.46 &  87.33 & \underline{90.43} \\
			ResNet$_{50}$ \cite{he2016deep}  & 92.31 & 90.00 & \textbf{96.81} \\
			\rc ResNet$_{101}$ \cite{he2016deep} & \textbf{94.23} & \textbf{94.00} & 84.04\\
			ViT$_{b/16}$\cite{dosovitskiy2020image} & \textbf{94.23} & \textbf{94.00} & 79.79 \\
			\rc ViT$_{b/32}$\cite{dosovitskiy2020image} & 90.38 & \textbf{94.00} & 78.72 \\
			\midrule
			Average & 91.92 & 91.87 & 85.96\\
			\bottomrule
		\end{tabularx}
            \vspace{-0.5em}
		\caption{Top-1 Accuracy for all Privacy and Action models on the original unmodified (source) videos, \ie without privacy obfuscation. 
  \textbf{Bold} and \underline{underline} indicate first and second best, resp. Number of frames and temporal sampling rate indicated as $f \times r$.}
		\label{tab:results_detail}
\end{table}

\subsection{Protocol}
\label{sec:protocol}
\noindent\textbf{Datasets.} We use three datasets commonly used for investigating action recognition and privacy, IPN \cite{ipn}, SBU \cite{sbu} and KTH \cite{kth}, which pose different challenges concerning both action and privacy. 

\noindent\underline{IPN} is a large-scale video-based hand gesture recognition dataset that consists of 50 actors performing 13 static or dynamic gestures, against three different backgrounds~\cite{ipn}. The performed gestures are used as the action labels and actor genders are used as the privacy labels.

\noindent\underline{SBU} is a video dataset depicting eight human interactions. 
Each video is a video of actor-pairs, in the same laboratory environment. Action labels are derived from the actor interactions and the privacy attributes are the unique pairings of seven different actors resulting in 13 privacy labels. 

\noindent\underline{KTH} has 25 actors doing one of six actions~\cite{kth}.
Each action is performed four times in different environments. The six action classes are used for action recognition and 25 actor identities serve as privacy labels. 
In the originally proposed splits, videos of any one actor are fully contained in one set, since the intent of the KTH dataset was not concerned with actor identity recognition. For privacy identity, each actor must appear in both sets; so, the data is split such that each action is performed twice by each person in the training set, and once in the validation and test sets.

\label{sec:metrics}
\noindent\textbf{Metrics.} Privacy recognition results are obtained following standard practice~\cite{bqn}, averaging outputs over multiple frames (32 for IPN and KTH, and 16 for SBU) from the same video. Action recognition results are obtained by reporting the top-1 accuracy, as per convention~\cite{priaction, prilens, pahmdb}.

It also is interesting to gauge the trade-off between action and privacy recognition. Let $0\leq a\leq 1$ 
be normalized action recognition performance derived from dividing accuracy percentages by $100$,
and let $p$ be defined analogously for privacy recognition. To quantify their trade-off we use a linear combination of $a$ and $1-p$ and define
\begin{align}
    f_{\lambda}(a,p) = (1-\lambda) a + \lambda (1-p),
    \label{eq:metric}
\end{align}
with $0\leq\lambda\leq 1$ weighing the relative importance of action recognition vs. privacy.
This metric is suitable to create a linear ranking as $a$, and $(1-p)$ are optimally $1$ ensuring $f_\lambda \in [0,1]$. 
If privacy or action recognition are not equally important, then they can be weighted accordingly; \eg if privacy is critical, then  $\lambda$ can be increased. In our main experiments we use $\lambda=0.5$, abbreviated as $f_{0.5}$ in Tab.~\ref{tab:results} and show an ablation over different values of $\lambda$ in Fig.~\ref{fig:averagemetricsplot}. 

\begin{figure}[!t]
	\centering
	\begin{tikzpicture}[spy using outlines={circle,yellow,magnification=2.5,size=1.8cm, connect spies}]
	\node {\pgfimage[interpolate=true,height=1.8cm]{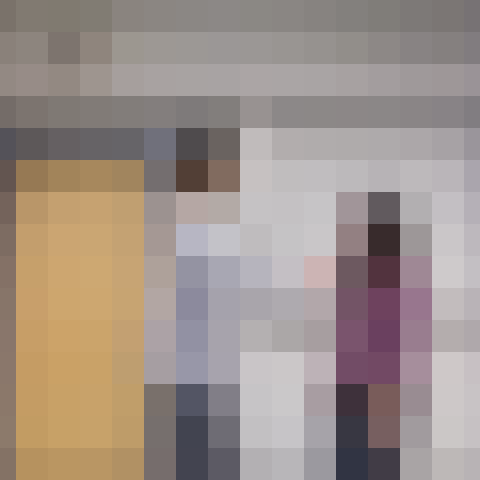}};
	\end{tikzpicture}%
	\begin{tikzpicture}[spy using outlines={circle,yellow,magnification=2.5,size=1.8cm, connect spies}]
		\node {\pgfimage[interpolate=true,height=1.8cm]{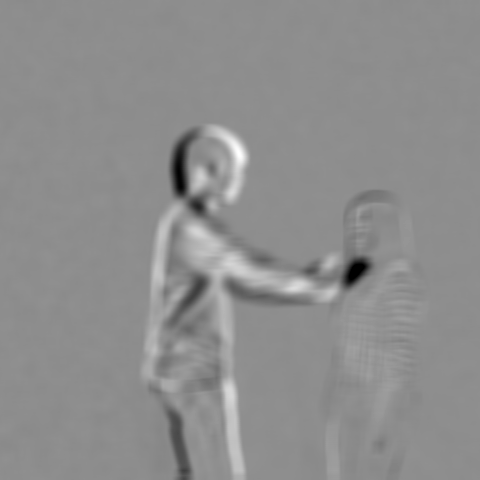}};
	\end{tikzpicture}%
	\begin{tikzpicture}[spy using outlines={circle,yellow,magnification=2.5,size=1.8cm, connect spies}]
		\node {\pgfimage[interpolate=true,height=1.8cm]{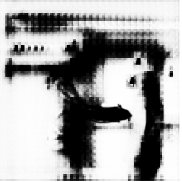}};
	\end{tikzpicture}%
	\begin{tikzpicture}[spy using outlines={circle,blue,magnification=2.5,size=1.8cm, connect spies}]
		\node {\pgfimage[interpolate=true,height=1.8cm]{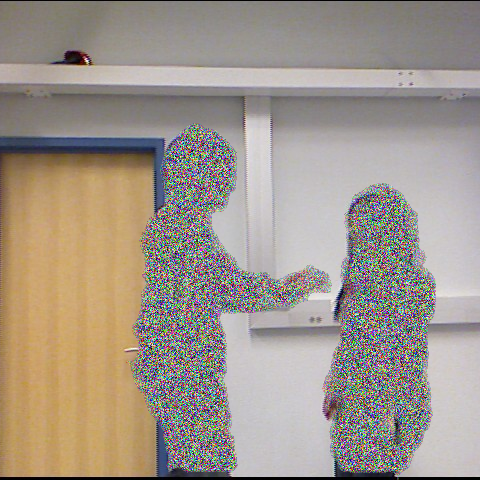}};
	\end{tikzpicture}
	
	\begin{tikzpicture}[spy using outlines={circle,blue!50,magnification=2.5,size=1.8cm, connect spies}]
		\node {\pgfimage[interpolate=true,height=1.8cm]{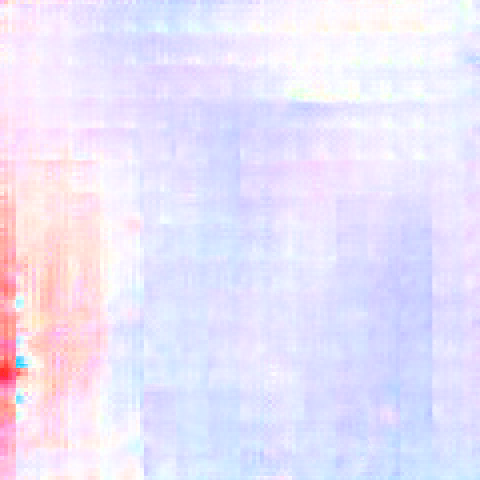}};
		\spy on (0.1, 0.1) in node [left] at (1,-2);
		\end{tikzpicture}%
		\begin{tikzpicture}[spy using outlines={circle,blue!50,magnification=2.5,size=1.8cm, connect spies}]
			\node {\pgfimage[interpolate=true,height=1.8cm]{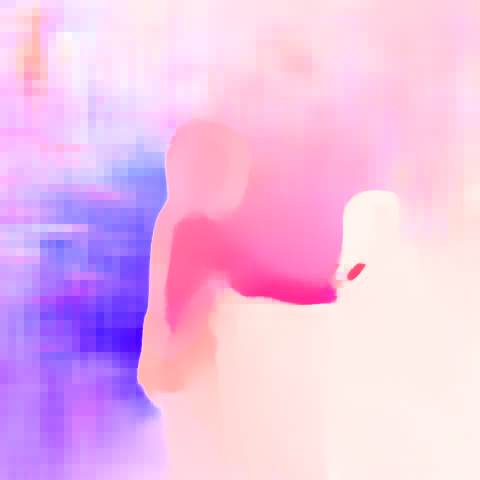}};
			\spy on (0.1, 0.1) in node [left] at (1,-2);
		\end{tikzpicture}%
		\begin{tikzpicture}[spy using outlines={circle,blue!50,magnification=2.5,size=1.8cm, connect spies}]
			\node {\pgfimage[interpolate=true,height=1.8cm]{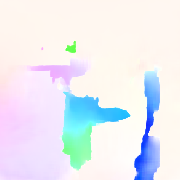}};
			\spy on (0.1, 0.1) in node [left] at (1,-2);
		\end{tikzpicture}%
		\begin{tikzpicture}[spy using outlines={circle, blue!50 ,magnification=2.5,size=1.8cm, connect spies}]
			\node {\pgfimage[interpolate=true,height=1.8cm]{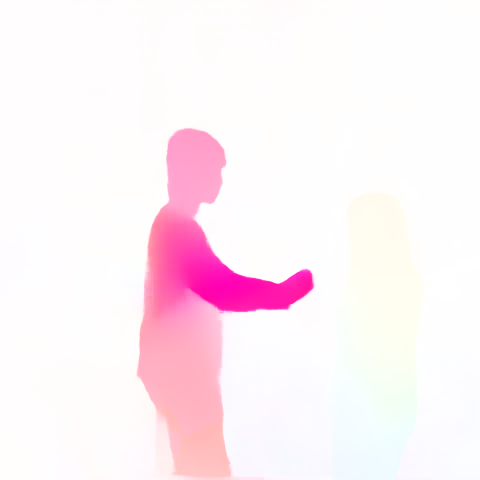}};
			\spy on (0.1, 0.1) in node [left] at (1,-2);
	\end{tikzpicture}
	\begin{tabularx}{\linewidth}{YYYY}
	\small{ELR~\cite{ryoo2018extremelowres}} & \small{BDQ~\cite{bqn}} & \small{ALF~\cite{pahmdb}} & \small{Ours} \\
	\end{tabularx}
	\vspace{-1.2em}
	\caption{Optical flow from two consecutive frames from competing approaches. Only our approach retains dynamic information that supports high-quality optical flow recovery. Top row: Single frames as processed by comparison algorithms. Middle row: Full frame optical flow recovery. Bottom row: Zoomed optical flow details. Optical flow shown in Middlebury colour coding~\cite{baker2011database}.}
	\label{fig:opticalflow}\vspace{-8pt}
\end{figure}

\vspace{0.5em}
\noindent\textbf{Competing obfuscation approaches.}
We briefly describe \textsc{Naive Baseline} approaches that solely rely on full-frame pixelation and bluring as well as actor masking that have been studied across the literature \cite{priaction,spact}. We also highlight other State-of-The-Art approaches (\textsc{SoTA}) \cite{ryoo2017privacy,pahmdb,bqn} that we compare against. Visual examples of the \textsc{SoTA} approaches are shown in Fig.~\ref{fig:opticalflow}; also shown are optical flows of consecutive frames for qualitative comparison.

\begin{table*}
	\small
	\centering
 	\textsc{Results across \underline{Naive Baseline} obfuscation methods}\vspace{0.2em}
		\begin{tabularx}{\linewidth}{lYYllYYllYYll}
			\toprule
			 & \multicolumn{4}{c}{IPN} & \multicolumn{4}{c}{KTH} & \multicolumn{4}{c}{SBU}\\ 
			\cmidrule(lr){2-4} \cmidrule(lr){6-8} \cmidrule(lr){10-12}
			\textbf{Method} &$\uparrow$Action&$\downarrow$Privacy& $\uparrow$$f_{0.5}$  \phantom{0}$\Delta$&& $\uparrow$Action&$\downarrow$Privacy& $\uparrow$$f_{0.5}$ \phantom{0}$\Delta$&& $\uparrow$Action&$\downarrow$Privacy& $\uparrow$$f_{0.5}$ \phantom{0}$\Delta$\\
			\midrule

		      Original 												& 88.05 &91.92  & 0.48 & & 90.15 & 91.87 & 0.49 & & 86.31 & 85.96 & 0.50 &\\
			\rc Masking 												& 36.45	&64.87	& 0.36 & & 37.54  &35.22 &0.51 & &52.62   &	30.1 & 0.61 &\\
			$\text{Pixelate}_{\tiny{4\times4}}$ 					& 85.59	&73.65	& \underline{0.56} && 83.64 &59.39 &\underline{0.62} &&73.24 &46.27 &0.63 &\\
			\rc $\text{Pixelate}_{\tiny{16\times16}}$ 					& 49.81 & 65.76 &0.42 & & 38.33 &38.43 &0.50 &&25.53&29.84&0.48&\\
			Blur $_{\tiny{\sigma=10}}^{\tiny{\kappa=13}} $ (weak)  	& 76.24 & 67.40	&0.54 & & 44.59 &37.73 &0.53 &&72.75 &35.69 &\underline{0.69}&\\
			\rc Blur $_{\tiny{\sigma=10}}^{\tiny{\kappa=21}} $ (strong) 	& 58.80	&65.92	&0.46 & & 30.38 &31.84 &0.49 &&57.77 &34.21 & 0.62&\\
			\rb Ours\phantom{paddinghereok} & 87.11 & 51.93 & \textbf{0.68} {\color{ForestGreen}{$+$\textbf{0.11}}}& &88.67 & \phantom{0}5.46 & \textbf{0.92} {\color{ForestGreen}{$+$\textbf{0.29}}}& & 86.74 & 13.19 & \textbf{0.87} {\color{ForestGreen}{$+$\textbf{0.18}}}& \\
			\bottomrule
		\end{tabularx}\\
\vspace{0.25em}\textsc{Results across \underline{SoTA} obfuscation methods}\vspace{0.01em}
		\begin{tabularx}{\linewidth}{lYYllYYllYYll}
			\midrule
			BDQ \cite{bqn}                   		      & 81.00 &59.00	&\underline{0.61} & &  91.11 &7.15 &\underline{0.92}&&84.04 &34.18 & \underline{0.75}&\\
			\rc ALF \cite{pahmdb}		       		  & 76.00	&65.00	&0.56 & & 85.89&19.27&0.83&&82.00 &48.00 &0.67 &\\
			ELR\cite{ryoo2018extremelowres} s=16		  & 70.82	&64.32	&0.53 & & 91.22 &88.86 &0.51 &&96.27 &82.97 &0.57 &\\
			\rc ELR\cite{ryoo2018extremelowres} s=32		  & 52.96	&63.29	&0.45 & & 85.57 &82.56 &0.52 &&92.42 &64.89 &0.64 &\\
			ELR\cite{ryoo2018extremelowres} s=64		  & 31.63	&62.70	&0.34 & & 56.21 &58.35 &0.49 &&80.05 &43.61 &0.68 &\\
			\rc Ours \textdagger & 85.25 & 51.67 & \underline{0.67} {\color{ForestGreen}{$+$\textbf{0.06}}}& & 89.44 & \phantom{0}4.31 & \textbf{0.93} {\color{ForestGreen}{$+$\textbf{0.01}}}& &	84.04 & 11.70 & \underline{0.86} {\color{ForestGreen}{$+$\textbf{0.11}}}&\\
			\rb Ours\phantom{paddinghereok} & 87.11 & 51.93 & \textbf{0.68} {\color{ForestGreen}{$+$\textbf{0.07}}}& &88.67 & \phantom{0}5.46 & \underline{0.92} {\color{Goldenrod}{$\pm$\textbf{0.00}}}& &86.74& 13.19 & \textbf{0.87} {\color{ForestGreen}{$+$\textbf{0.12}}}& \\

			\bottomrule
		\end{tabularx}
		\vspace{-0.5em}
		\caption{Comparison between Ours vs.~\textsc{Naive Baseline} (top) and \textsc{SoTA} (bottom) Approaches as Top-1 Accuracy for both Action and Privacy labels, as well as $f_{0.5}$ Introduced in Sec.~\ref{sec:protocol}. We show Ours and \textsc{Naive Baseline} results averaged across all recognition algorithms presented in Tab.~\ref{tab:results_detail}. Competing \textsc{SoTA} approaches only present results on one specific algorithm for action recognition (I3D \cite{carreira2017quo}) and privacy attribute detection (ResNet50 \cite{he2016deep}) as those approaches are network specific. To showcase a fair comparison we also present results with the same single recognition algorithms indicated as  ``Ours \textdagger". First and second best results indicated by \textbf{bold} and \underline{underline}, respectively. Relative performance delta ($\Delta$) is indicated in {\color{ForestGreen}{\textbf{green}}} (improvement) and {\color{Goldenrod}{\textbf{orange}}} (tie) between best and second best method.
        }
		\label{tab:results}
\vspace{-1.5em}
\end{table*}

\vspace{0.2em}
\noindent\underline{Mask Obfuscation}. All our datasets involve people performing actions and the privacy attributes are related to people. A naive way to preserve privacy in such videos is to completely mask out the actors. To do so, we use the YOLOv8 implementation \cite{yolov8} based on the original YOLO \cite{redmon2016you}. We report results based on the masked region filled with the mean intensity of the image covered by the mask. 

\vspace{0.2em}
\noindent\underline{Pixelation}. Pixelation applies average pooling over regions of dimensions $x \times x$ on the input image sequence. We choose  two scales of $ x \in \{4,16\}$ for the patch-size to pool. In the result tables these methods are abbreviated as Pix$_{x \times x}$.
    
\vspace{0.2em}
\noindent\underline{Blur}. 
Blur applies a Gaussian blur to the images that have been rescaled to $224 \times 224$ pixels. The parameters of the Gaussian are the kernel size, $\kappa$, and the standard deviation of the kernel, $\sigma$. We choose values for weak and strong blurs, $_{\tiny{\sigma=10}}^{\tiny{\kappa=13}} $ and $_{\tiny{\sigma=10}}^{\tiny{\kappa=21}}$, respectively, as consistent with other work on obfuscation methods \cite{pahmdb}.

\vspace{0.2em}
\noindent\underline{ELR} initially reduces frames to Extreme Low Resolution and subsequently applies a set of learned inverse super-resolution transforms to 
support action recognition~\cite{ryoo2017privacy}. 

\vspace{0.2em}
\noindent\underline{ALF} is an Adversarial Learning Framework for action recognition that takes into account a privacy budget~\cite{pahmdb}. 

\vspace{0.2em}
\noindent\underline{BDQ} is a privacy-preserving encoder that sequentially Blurs, Differences and Quantizes frames. The blur and quantization parameters are learned to maximize action recognition while minimizing privacy recognition~\cite{bqn}. 

\vspace{0.5em}
\noindent Notably, all the compared \textsc{SoTA} methods operate across entire frames (\ie without selectivity) and have limited interpretability due to their learning-based obfuscation.


\vspace{0.5em}
\noindent\textbf{Implementation details.}

\noindent\underline{Training}.
No large scale dataset exists that allows for joint training of action and privacy classification on the same videos. Therefore, we pretrain our action and privacy networks on Kinetics400~\cite{carreira2017quo} and Imagenet1k~\cite{imagenet}, respectively, and then finetune for specific datasets. We use the AdamW~\cite{adam} optimizer with a learning rate of $3e^{-4}$. The networks are trained with a patience scheme of 100 epochs that monitors the loss on the validation set. 
For action recognition, videos are temporally uniformly subsampled according to the architecture (Tab.~\ref{tab:results_detail}, `$f \times r$' column). 

Note that we never perform any training on videos obfuscated by our approach. To implement warping, \eqref{eq:rwarp}, we use pretrained RAFT to extract optical flow~\cite{RAFT}.

\vspace{0.2em}
\noindent\underline{Privacy template selection}. 
Our approach affords selective combination of a predefined set of privacy templates for a given dataset or application. For the experiments, we choose a subset of templates, $\tilde{\mathsf{T}}$, from our complete identity preserving template library, $\mathsf{T}$, shown in Fig.~\ref{fig:templates}, to optimize performance on a given dataset. Figure~\ref{fig:individual} shows template-wise performance based on both action recognition and privacy for each dataset. For SBU and KTH, action recognition is stable with respect to templates; however, the same four templates notably reduce privacy recognition. For consistency on IPN, we also select the four templates that most reduce privacy recognition, even while preserving action recognition; although, the distinction is less striking.
This selection process results in templates $\tilde{\mathsf{T}} = \{ \mathit{torso},\mathit{arm},\mathit{leg},\mathit{hair}\}$ for SBU and KTH vs. $\tilde{\mathsf{T}} = \{ \mathit{cheek},\mathit{eyes},\mathit{forehead},\mathit{hair}\}$ for IPN. 
Our code yields more results on the overlap of individual saliency maps.

\noindent\underline{Runtime}.
The constant overhead to produce our obfuscated videos is 
$\approx$$100ms$/frame on a NVIDIA RTX4080 GPU.

\subsection{Results}
\label{sec:results}

We evaluate our approach to privacy preserving action recognition on 13 different action recognition models and five different privacy attribute recognition models. 
Competing approaches evaluate on at most five action models \cite{ryoo2017privacy,pahmdb,bqn}.
Table~\ref{tab:results_detail} shows results 
on the original videos from IPN \cite{ipn}, SBU \cite{sbu} and KTH \cite{kth}, \ie without any privacy preserving processing. 
Table~\ref{tab:results} shows results comparing our approach vs.~\textsc{Naive Baselines} and \textsc{SoTA}. Comparison is given as the average across recognition algorithms presented in Table~\ref{tab:results_detail}. 
Notably, while our approach compromises at most $1.48$\% action recognition accuracy compared to performance on the original video ($88.67$\% vs.~$90.15$\% on KTH), it \textit{always} greatly improves privacy.

\vspace{0.2em}
\noindent\textbf{Naive Baselines.} 
Compared to the baselines, our approach scores highest with respect to the $f_{0.5}$ metric across all three datasets: IPN ({\color{ForestGreen}{$+$\textbf{0.11}}}), KTH ({\color{ForestGreen}{$+$\textbf{0.29}}}), and  SBU ({\color{ForestGreen}{$+$\textbf{0.18}}}). 
\textsc{Naive Baselines} show that action recognition correlates highly with privacy performance, as $f_{0.5}$ hovers around 0.5. The notable exception is $\text{Pix}_{\tiny{4\times4}}$ that performs well on IPN and KTH, and is the second best \textsc{Naive Baseline} on SBU. A notable benefit of all these approaches is that they do not require retraining of the recognition algorithms. Still, none have a selective ability to obfuscate only certain parts. 

\vspace{0.2em}
\noindent\textbf{State of the art.} A noteworthy difference between our vs. the alternative \textsc{SoTA} approaches is that they only work with the recognition algorithms on which they were trained. While this fact disadvantages our approach, as it lacks such retraining, it still outperforms the competing methods on all datasets: IPN ({\color{ForestGreen}{$+$\textbf{0.06}}}), KTH ({\color{ForestGreen}{$+$\textbf{0.01}}}), and  SBU ({\color{ForestGreen}{$+$\textbf{0.11}}}). 

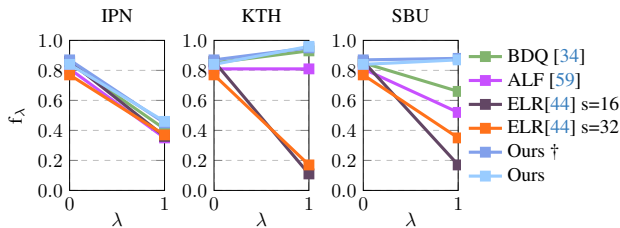
\begin{figure}[!t]
	\centering
	\resizebox{1.\linewidth}{!}{%
	\begin{tikzpicture}
		\begin{axis}[
			title={IPN},
                xlabel=$\lambda$,
                ylabel=f$_\lambda$,
			xmin=0, xmax=1.0,
			ymin=0, ymax=1.0,
                xtick={0.0,1.0},
			ytick={0.0,0.20,0.40,0.60,0.80,1.0,1.2},
			legend pos=outer north east,
			ymajorgrids=true,
			xmajorgrids=true,
			grid style=dashed,
			colormap name=bright, 
			cycle list={[of colormap]},
			every axis plot/.append style={mark=square*,ultra thick},
			legend cell align={left},
			width=3.3cm,
			height=4.3cm,
               xlabel style={at={(axis description cs:0.5,-0.1)},anchor=north},
			yticklabel style={/pgf/number format/.cd,
            fixed,
            fixed zerofill,
            precision=1,
			/tikz/.cd},
			xticklabel style={/pgf/number format/.cd,
            fixed,
            fixed zerofill,
            precision=0,
			/tikz/.cd},
			]
		]
                \addplot+[]
                coordinates {
                        (0.00, 0.85)(1.00, 0.41)};

                \addplot+[]
                coordinates {
                        (0.00, 0.81)(1.00, 0.35)};

                \addplot+[]
                coordinates {
                        (0.00, 0.86)(1.00, 0.36)};

                \addplot+[]
                coordinates {
                        (0.00, 0.77)(1.00, 0.37)};

                \addplot+[style={mark=square*,ultra thick}]
                coordinates {
                        (0.00, 0.87)(1.00, 0.45)};

                \addplot+[style={mark=square*,ultra thick}]
                coordinates {
                        (0.00, 0.84)(1.00, 0.46)};
		\end{axis}
	\end{tikzpicture}%
 	\begin{tikzpicture}
		\begin{axis}[
			title={KTH},
                xlabel=$\lambda$,
			xmin=0, xmax=1.0,
			ymin=0, ymax=1.0,
                xtick={0.0,1.0},
			ytick={0.0,0.20,0.40,0.60,0.80,1.0,1.2},
			legend pos=outer north east,
			ymajorgrids=true,
			xmajorgrids=true,
			grid style=dashed,
			colormap name=bright, 
			cycle list={[of colormap]},
			every axis plot/.append style={mark=square*,ultra thick},
			legend cell align={left},
			width=3.3cm,
			height=4.3cm,
                xlabel style={at={(axis description cs:0.5,-0.1)},anchor=north},
			yticklabel style={/pgf/number format/.cd,
            fixed,
            fixed zerofill,
            precision=1,
			/tikz/.cd},
			xticklabel style={/pgf/number format/.cd,
            fixed,
            fixed zerofill,
            precision=0,
			/tikz/.cd},
			]
		]
                \addplot+[]
                coordinates {
                        (0.00, 0.85)(1.00, 0.93)};

                \addplot+[]
                coordinates {
                        (0.00, 0.81)(1.00, 0.81)};

                \addplot+[]
                coordinates {
                        (0.00, 0.86)(1.00, 0.11)};

                \addplot+[]
                coordinates {
                        (0.00, 0.77)(1.00, 0.17)};

                \addplot+[style={mark=square*,ultra thick}]
                coordinates {
                        (0.00, 0.87)(1.00, 0.95)};

                \addplot+[style={mark=square*,ultra thick}]
                coordinates {
                        (0.00, 0.84)(1.00, 0.96)};
		\end{axis}
	\end{tikzpicture}	
 	\begin{tikzpicture}
		\begin{axis}[
			title={SBU},
                xlabel=$\lambda$,
			xmin=0, xmax=1.0,
			ymin=0, ymax=1.0,
                xtick={0.0,1.0},
			legend pos=outer north east,
			ymajorgrids=true,
			xmajorgrids=true,
			grid style=dashed,
			colormap name=bright, 
			cycle list={[of colormap]},
			every axis plot/.append style={mark=square*,ultra thick},
			legend cell align={left},
			width=3.3cm,
			height=4.3cm,
               legend style={draw=none},
               xlabel style={at={(axis description cs:0.5,-0.1)},anchor=north},
			yticklabel style={/pgf/number format/.cd,
            fixed,
            fixed zerofill,
            precision=1,
			/tikz/.cd},
			xticklabel style={/pgf/number format/.cd,
            fixed,
            fixed zerofill,
            precision=0,
			/tikz/.cd},
			]
		]
                \addplot+[]
                coordinates {
                        (0.00, 0.85)(1.00, 0.66)};
                \addlegendentry{BDQ \cite{bqn}}

                \addplot+[]
                coordinates {
                        (0.00, 0.81)(1.00, 0.52)};
                \addlegendentry{ALF \cite{pahmdb}}

                \addplot+[]
                coordinates {
                        (0.00, 0.86)(1.00, 0.17)};
                \addlegendentry{ELR\cite{ryoo2018extremelowres} s=16}

                \addplot+[]
                coordinates {
                        (0.00, 0.77)(1.00, 0.35)};
                \addlegendentry{ELR\cite{ryoo2018extremelowres} s=32}

                \addplot+[style={mark=square*,ultra thick}]
                coordinates {
                        (0.00, 0.87)(1.00, 0.88)};
                \addlegendentry{Ours \textdagger}

                \addplot+[style={mark=square*,ultra thick}]
                coordinates {
                        (0.00, 0.84)(1.00, 0.87)};
                \addlegendentry{Ours}

		\end{axis}
	\end{tikzpicture}%
 }%
        \vspace{-1.0em}
	\caption{Performance across all datasets with varying $\lambda$;  see \eqref{eq:metric}. Values of $\lambda$ closer to $0$ weigh action recognition higher, whereas values closer to $1$ increase the importance of privacy preservation.
 }
 \vspace{-1.4em}
\label{fig:averagemetricsplot}
\end{figure}

It is valuable to consider obfuscation approaches in terms of the relative importance of action recognition vs. privacy preservation. Figure~\ref{fig:averagemetricsplot} compares results for the \textsc{SoTA} approaches as that trade-off is varied in terms of $f_{\lambda}$, \eqref{eq:metric}. The sweep of $\lambda$ shows that our obfuscation approach performs better than any competing approach across the entire range. Especially with increasing $\lambda$, the gap to other approaches widens as more emphasis is put on privacy preservation.

\vspace{0.2em}
\noindent\textbf{Importance of selective obfuscation.}
Our approach is unique in its ability to selectively obfuscate particular regions within a video based on specific privacy attributes. This ability yields two benefits: 

(i) Selectivity aids in interpretability.
Each selected template results in a saliency map, \eqref{eq:saliencyMap}, which allows for visual inspection of what information is being obscured.
Figure~\ref{fig:saliency} highlights this benefit as the heat maps reveal the degree of obfuscation to be applied on a template-by-template basis. 

(ii) Individual privacy templates can be chosen and combined for best performance; see Fig.~\ref{fig:individual}. Depending on the dataset, different privacy templates differently impact the performance of privacy and action recognition. Action recognition on SBU and KTH is relatively robust to privacy template selection; however, privacy preservation is best when a subset of templates (\textit{torso}, \textit{leg}, \textit{arm}, \textit{hair}) is selected and the rest remain unobfuscated.
This fact derives directly from the saliency calculation, \eqref{eq:saliency}: If multiple individual saliency maps have small values, 
then other strong responses are scaled down in the combined final saliency map.
Subsequently, this effect leads to less noise being applied in our selective obfuscation, \eqref{eq:rselective}, which can deteriorate privacy preservation.
In contrast, action recognition on IPN is compromised if the \textit{hand} template is selected (as expected with a gesture centric dataset), which documents that simply obfuscating the entire person leads to inferior action performance compared to selectively applied obfuscation.


\vspace{0.2em}
\noindent\textbf{Importance of motion consistent noise.}
\label{sec:motionnoise}
Action recognition is complicated as different datasets, and even individual actions within a dataset might require different recognition capabilities and can rely to different degrees on the modeling of motion~\cite{choi2019can}.
This uncertainty on the role of motion in action recognition is exacerbated by the fact that different deep learning-based architectures are successful in capturing motion to varying degrees~\cite{kowal2022deeper}. 

\begin{figure}[!t]
	\centering
	\resizebox{0.5\linewidth}{!}{%
	\begin{tikzpicture}
		\begin{axis}[
			title={\Large{Action}},
			ylabel={Top-1 Accuracy},
			ybar=2*\pgflinewidth,
			bar width=20pt,
			symbolic x coords={IPN,KTH,SBU},
			enlarge x limits=0.25,
			xtick = data,
			ytick={0,20,40,60,80,100},
			ymin=0, ymax=100,
			ymajorgrids=true,
			grid style=dashed,
			legend pos=south west,
			legend cell align={left},
			tick label style={font=\Large},
			label style={font=\Large},
			height = 5cm,
			width = 8cm,
		]
			\addplot+[]
				coordinates {(IPN, 87.11) (KTH,88.67) (SBU,86.74)};
				\addlegendentry{\Large{Ours}}

			\addplot+[]
				 coordinates {(IPN,84.40) (KTH,59.12) (SBU,68.30)};
				 \addlegendentry{\Large{$iid$ Noise}}

		\end{axis}
	\end{tikzpicture}
	}%
	\resizebox{0.5\linewidth}{!}{%
	\begin{tikzpicture}
		\begin{axis}[
			title={\Large{Privacy}},
			ylabel={\phantom{Top-1 Accuracy}},
			ybar=2*\pgflinewidth,
			bar width=20pt,
			symbolic x coords={IPN,KTH,SBU},
			enlarge x limits=0.25,
			xtick = data,
			ytick={0,20,40,60,80,100},
			ymin=0, ymax=100,
			ymajorgrids=true,
			grid style=dashed,
			legend pos=north east,
			legend cell align={left},
			tick label style={font=\Large},
			label style={font=\Large},
			height = 5cm,
			width = 8cm,
		]
			\addplot+[]
			coordinates {(IPN, 51.93) (KTH,5.46) (SBU,13.19)};
			\addlegendentry{\Large{Ours}}

			\addplot+[]
			coordinates {(IPN, 51.93) (KTH,5.46) (SBU,13.19)};
			\addlegendentry{\Large{$iid$ Noise}}

		\end{axis}
	\end{tikzpicture}
	}%
\vspace{-0.5em}
\caption{
Comparing Temporally Consistent vs.\ $iid$ Obfuscation for Action Recognition and Privacy Preservation. Action recognition performance decreases if noise is not temporally consistent.
}
\vspace{-1em}
\label{fig:iidablation}
\end{figure}
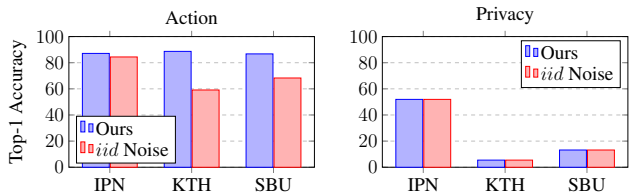

To see the importance of our proposed motion consistent noise for privacy preservation while maintaining good action recognition, 
we compare to another version of our pipeline that obfuscates video frames using independent, identically distributed ($iid$) noise; see Fig.~\ref{fig:iidablation} and project page for video results.
For all datasets, action recognition performance decreases if the noise is $iid$. This result is sensible, because the $iid$ noise does not capture the temporal dynamics of the source video.
In contrast, privacy preservation is robust to noise in either case. There is less impact on IPN action recognition, as the critical hand motion is never obfuscated, which underlines the importance of selective masking. Further insight on why motion consistent noise supports better action can be had through consideration of Fig.~\ref{fig:opticalflow}, where the ability of such noise to support optical flow estimation is illustrated. 


\vspace{0.2em}
\noindent\textbf{Limitations.}\label{sec:motionnoise}
Inevitably, there will be a trade-off between action recognition and
privacy, as relevant information may be shared. In
our approach, that trade-off could happen 
because our temporally consistent
noise maintains dynamic information important for action recognition; however, it also might support motion based identification, \eg gait recognition. Current protocols in privacy preserving action recognition do not consider motion-based identification. However, gait and related motion-based measurements are weak biometrics~\cite{jain2007handbook}; so, favouring action recognition may be apt. Moreover, if it becomes a concern, then it may be possible to apply motion perturbations that impede personal identification while maintaining action recognition.
\vspace{-0.2em}
\section{Conclusion}
\vspace{-0.1em}
Our work highlights that it is not necessary to train action recognition and privacy networks in an adversarial fashion for effective obfuscation of privacy attributes while maintaining strong action recognition performance. We show that a system based on local privacy templates, deep features that capture template semantics and selective noise obfuscation that is animated with source video motion can uphold privacy without hindering action recognition.
Our approach is unique compared to alternative recent approaches in terms of interpretability and independence from particular action and privacy recognition algorithms. Our nine manually chosen templates, in combination with our proposed obfuscation technique outperforms other state-of-the art approaches across three different datasets. 



\cleardoublepage
{
    \small
    \bibliographystyle{ieeenat_fullname}
    \bibliography{main}
}


\end{document}